\documentclass{article}
\usepackage{amsmath,amssymb}
\usepackage{graphicx}
\usepackage{hyperref}

\newcommand{\T}{^{\text{\tiny\sffamily\upshape\mdseries T}}}

\PassOptionsToPackage{numbers, sort, compress}{natbib}

\usepackage[preprint]{neurips_2024}

\usepackage[utf8]{inputenc} 
\usepackage[T1]{fontenc}    
\usepackage{hyperref}       
\usepackage{url}            
\usepackage{booktabs}       
\usepackage{amsfonts}       
\usepackage{nicefrac}       
\usepackage{microtype}      
\usepackage{xcolor}         

\usepackage{subcaption}
\usepackage{graphicx}
\usepackage{multirow}
\usepackage{amsmath,amssymb,amsfonts}
\usepackage{amsthm}
\usepackage{mathrsfs}
\usepackage{xcolor}
\usepackage{textcomp}
\usepackage{manyfoot}
\usepackage{booktabs}
\usepackage{algorithm}
\usepackage{algorithmicx}
\usepackage{algpseudocode}
\usepackage{listings}

\newtheorem{theorem}{Theorem} 
\newtheorem{lemma}{Lemma}%

\newtheorem{assumption}{Assumption}

\title{Unraveling the Hessian: A Key to Smooth Convergence in Loss Function Landscapes}

\author{%
  Nikita Kiselev\\
  MIPT, Sber AI\\
  Moscow, Russia\\
  \texttt{kiselev.ns@phystech.edu}\\
  \And
  Andrey Grabovoy\\
  MIPT\\
  Moscow, Russia\\
  \texttt{grabovoy.av@phystech.edu}\\
}

\begin{document}

\maketitle

\begin{abstract}
    The loss landscape of neural networks is a critical aspect of their training, and understanding its properties is essential for improving their performance. In this paper, we investigate how the loss surface changes when the sample size increases, a previously unexplored issue. We theoretically analyze the convergence of the loss landscape in a fully connected neural network and derive upper bounds for the difference in loss function values when adding a new object to the sample. Our empirical study confirms these results on various datasets, demonstrating the convergence of the loss function surface for image classification tasks. Our findings provide insights into the local geometry of neural loss landscapes and have implications for the development of sample size determination techniques.
\end{abstract}


\section{Introduction}\label{sec:intro}

\begin{figure}[ht]
    \centering
    \includegraphics[width=0.65\linewidth]{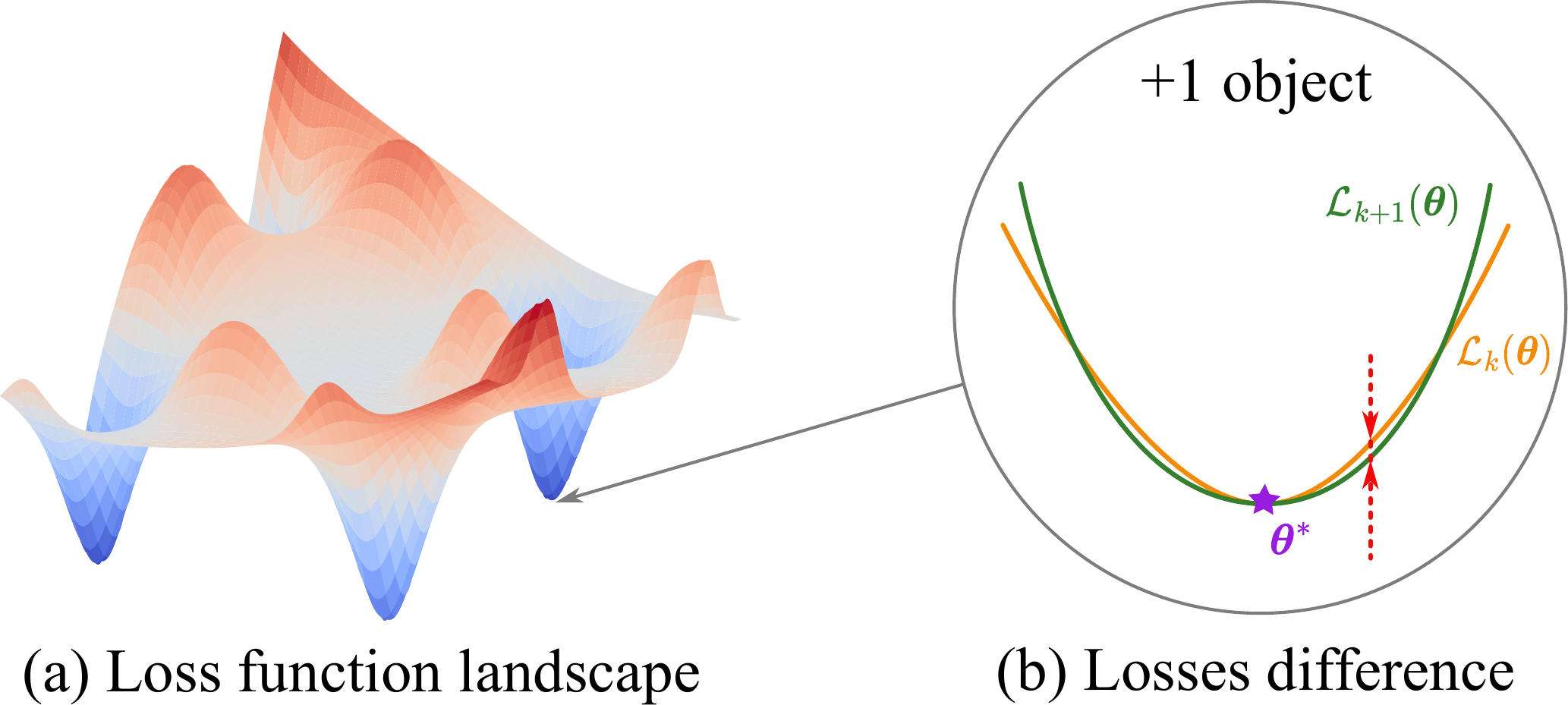}
    \caption{\textbf{Overview of our observations.} Part (a) shows the loss function landscape, which is a surface in the parameters space. Part (b) shows the losses difference. It arises, when one more object is added to the dataset. Here we exhibit the behavior for dimension equals 2. Near the minimum $\boldsymbol{\theta}^*$, the mean loss value for $k+1$ objects $\mathcal{L}_{k+1}(\boldsymbol{\theta})$ tends to be similar to the same for $k$ objects $\mathcal{L}_{k}(\boldsymbol{\theta})$.}
    \label{fig:overview}
\end{figure}

The advancement of neural networks has significantly driven the development of optimization methods \cite{choi2020empiricalcomparisonsoptimizersdeep, Soydaner_2020, schmidt2021descendingcrowdedvalley}. Nevertheless, a notable challenge lies in the extensive number of parameters, which can result in numerous potential local and global minima of the loss function \cite{neyshabur2018the, zou2019improvedanalysistrainingoverparameterized, allenzhu2019convergencetheorydeeplearning, allenzhu2020learninggeneralizationoverparameterizedneural}. Many studies have endeavored to identify the most optimal minima from diverse perspectives \cite{choromanska2015losssurfacesmultilayernetworks, you2017largebatchtrainingconvolutional, li2018visualizinglosslandscapeneural}.

Flatter minima are associated with superior generalization properties. This issue has been explored theoretically and empirically in many studies, including \cite{NIPS1994_01882513, neyshabur2017exploringgeneralizationdeeplearning, dinh2017sharpminimageneralizedeep, fort2019emergentpropertieslocalgeometry}. The process of locating a flat minimum is not straightforward. In such contexts, the Hessian matrix, which represents the second-order derivative of the loss function, is frequently utilized. This matrix is crucial for understanding the local behavior of the function around the point of extremum.

Existing research \cite{pmlr-v97-ghorbani19b, sagun2018empiricalanalysishessianoverparametrized, papyan2019spectrumdeepnethessiansscale, liao2021hessianeigenspectrarealisticnonlinear, dauphin2024neglectedhessiancomponentexplains, papyan2020tracesclasscrossclassstructurepervade, papyan2019measurementsthreelevelhierarchicalstructure, garrod2024unifyinglowdimensionalobservations} has addressed the question of Hessian spectra in typical neural network architectures, revealing a characteristic pattern. The spectrum typically comprises a significant number of eigenvalues close to zero, referred to as bulk eigenvalues, and a smaller number of distinct eigenvalues, known as outliers \cite{pmlr-v97-ghorbani19b}. Specifically, in a $K$-label classification task, precisely $K$ non-zero eigenvalues can be observed \cite{sagun2018empiricalanalysishessianoverparametrized, papyan2019spectrumdeepnethessiansscale}. Theoretical estimations of these non-vanishing eigenvalues enable the establishment of an upper bound on the local rate of growth of the loss function.

However, the issue of altering the landscape of the loss function when adding new objects to the sample remains unresolved. In this paper, we meticulously investigate how the loss surface transforms as the sample size increases. Specifically, we examine the absolute value of the loss function changes when adding another object. An overview of our observations is presented in Figure~\ref{fig:overview}.

We obtain theoretical estimates for the convergence of this difference in a fully connected neural network as the sample size approaches infinity. These results are derived through the analysis of the Hessian spectrum. These estimates allow us to determine the dependence of this difference on the structure of the neural network, including the size of the hidden layer and the number of layers. We empirically verify these theoretical results by examining the behavior of the loss surface on various datasets. The obtained plots substantiate the validity of the theoretical calculations.

\textbf{Contributions.} Our contributions can be summarized as follows:
\begin{itemize}
    \item We present a theoretical analysis of the convergence of the loss landscape in a fully connected neural network as the sample size increases, deriving upper bounds for the difference in loss function values when adding a new object to the sample.
    \item We demonstrate the validity of our theoretical results through empirical studies on various datasets, showing that the loss function surface exhibits convergence for image classification tasks.
    \item We highlight the implications of our findings for understanding the local geometry of neural loss landscapes and for the development of sample size determination techniques, addressing a previously unexplored issue in the field.
\end{itemize}

\textbf{Outline.} The rest of the paper is organized as follows. Section~\ref{sec:rw} divides existing results into some topics, highlighting their key contributions and findings. Section~\ref{sec:prelim} considers general notation and some preliminary calculations. In Section~\ref{sec:converg}, we provide theoretical bounds on the hessian and losses difference norms. Empirical study of the obtained results is given in Section~\ref{sec:exp}. We summarize and present the results in Sections~\ref{sec:disc} and~\ref{sec:concl}. Additional experiments and proofs of theorems are included in the Appendix~\ref{app}.

\section{Related Work}\label{sec:rw}

\textbf{Understanding the Geometry of Neural Network Loss Landscapes.}
The geometry of neural network loss landscapes, particularly through the Hessian matrix, has been extensively studied. \cite{fort2019emergentpropertieslocalgeometry} identifies key properties including that in the multi-label classificaion problem the landscape exhibits exactly $K$ directions of high positive curvature, where $K$ is the number of classes. \cite{pmlr-v70-pennington17a} and \cite{pmlr-v97-ghorbani19b} use random matrix theory and spectral analysis to understand the loss surface dynamics and optimization. \cite{singh2024landscapinglinearmodeconnectivity} introduces a model for understanding linear mode connectivity (LMC) by analyzing the topography of the loss landscape. \cite{singh2022phenomenologydoubledescentfinitewidth} provides a theoretical understanding of the double descent phenomenon in finite-width neural networks, leveraging influence functions to derive expressions for the population loss. \cite{wang2023instabilitieslargelearningrate} characterizes the instabilities of gradient descent during training with large learning rates, observing landscape flattening and shift. However, nowhere is the question raised that this geometry becomes unchanged with an increase in the number of objects.

\textbf{Hessian-Based Generalization and Optimization.}
The Hessian matrix is crucial for studying generalization and optimization in neural networks. In \cite{ju2023robustfinetuningdeepneural} a Hessian-based distance for fine-tuning against label noise was introduced, which can match the scale of the observed generalization gap of fine-tuned models in practice. \cite{nguyen2024agnosticsharpnessawareminimization} optimizes for wider local minima using training data, while concurrently maintaining low loss values on validation data to improve generalization. Authors of \cite{macdonald2023progressivesharpeningflatminima} ties loss curvature to input-output behavior, explaining progressive sharpening and flat minima. All these studies provide insights into optimizing and generalizing neural networks using Hessian-based methods. However, none of these papers address the impact of changing sample size.

\textbf{Spectral Analysis and Structural Insights.}
Spectral analysis of the Hessian matrix offers insights into neural network structure and properties. There is a variety of works \cite{sagun2018empiricalanalysishessianoverparametrized,pmlr-v97-ghorbani19b,papyan2019spectrumdeepnethessiansscale} underlying that the Hessian matrix of typical loss surface exhibits a spectrum composed of two parts: a bulk centered near zero and outliers away from the bulk. \cite{liao2021hessianeigenspectrarealisticnonlinear} shows that depending on the data properties, the nonlinear response model, and the loss function, the Hessian can have qualitatively different spectral behaviors. \cite{dauphin2024neglectedhessiancomponentexplains} highlights the Hessian's role in sharpness regularization. \cite{papyan2020tracesclasscrossclassstructurepervade} reveals class/cross-class structure in Hessian spectra. \cite{papyan2019spectrumdeepnethessiansscale} and \cite{papyan2019measurementsthreelevelhierarchicalstructure} analyze Hessian dynamics and hierarchical structure. \cite{garrod2024unifyinglowdimensionalobservations} unifies low-dimensional observations, explaining Hessian spectra and gradient descent alignment. \cite{sagun2017eigenvalueshessiandeeplearning} and \cite{sagun2018empiricalanalysishessianoverparametrized} analyze the Hessian's eigenvalue distribution, highlighting over-parametrization and data dependency. \cite{xie2022powerlawhessianspectrumsdeep} discovers and mathematically models the power-law Hessian spectrum, providing a maximum entropy interpretation and a framework for spectral analysis in deep learning. However, these low-rank approximations have not been sufficiently investigated in terms of the convergence of the corresponding spectra.

\textbf{Decomposing and Analyzing the Hessian Matrix.}
Decomposing the Hessian matrix provides insights into neural network training and generalization. \cite{wu2022dissectinghessianunderstandingcommon} proposes a decoupling conjecture to analyze Hessian properties, decomposing it as the Kronecker product of two smaller matrices. \cite{singh2023hessianperspectivenatureconvolutional} explores CNN Hessian maps, revealing the Hessian rank grows as the square root of the number of parameters. \cite{singh2021analyticinsightsstructurerank} provides exact formulas and tight upper bounds for the Hessian rank of deep linear networks. \cite{skorski2019chainruleshessianhigher} simplifies high-dimensional derivative calculations using tensor calculus. However, the limit properties of the Hessian with increasing sample size have not been sufficiently investigated. 

\section{Preliminaries}\label{sec:prelim}

\subsection{General notation}

In this section, we introduce the general notation used in the rest of the paper and the basic assumptions. Consider a conditional probability $p(\mathbf{y}|\mathbf{x})$, that maps the given unobserved variable $\mathbf{x} \in \mathcal{X}$ to the corresponding output $\mathbf{y} \in \mathcal{Y}$, and which we try to approximate using neural network $f_{\boldsymbol{\theta}}(\cdot)$ with parameters $\boldsymbol{\theta} \in \mathbb{R}^{P}$.

Given the dataset
\[ \mathfrak{D} = \left\{ (\mathbf{x}_i, \mathbf{y}_i) \right\}, \quad i = 1, \ldots, m, \]
the empirical loss function calculated for all the given dataset of size $m$ is
\[ \mathcal{L}(\boldsymbol{\theta}) = \dfrac{1}{m} \sum\limits_{i=1}^{m} \ell(f_{\boldsymbol{\theta}}(\mathbf{x}_i), \mathbf{y}_i) \approx \mathbb{E}_{(\mathbf{x}, \mathbf{y}) \sim p(\mathbf{x}, \mathbf{y})} \left[ \ell(f_{\boldsymbol{\theta}}(\mathbf{x}), \mathbf{y}) \right], \]
so it is an approximation of general loss function, which is calculated using the joint distribution $p(\mathbf{x}, \mathbf{y})$. Here we denote the per-object loss function as $\ell(\mathbf{z}, \mathbf{y})$, e.g., cross-entropy loss $\mathrm{CE}(\mathrm{softmax}(\mathbf{z}), \mathbf{y})$ in multi-label classification. If we fix first $k$ samples, corresponding loss function is
\[ \mathcal{L}_k(\boldsymbol{\theta}) = \dfrac{1}{k} \sum\limits_{i=1}^{k} \ell(f_{\boldsymbol{\theta}}(\mathbf{x}_i), \mathbf{y}_i). \]
Difference between losses for sample sizes $k+1$ and $k$ is
\begin{equation}\label{eq:difference}
    \mathcal{L}_{k+1}(\boldsymbol{\theta}) - \mathcal{L}_k(\boldsymbol{\theta}) = \dfrac{1}{k+1} \left( \ell(f_{\boldsymbol{\theta}}(\mathbf{x}_{k+1}), \mathbf{y}_{k+1}) - \mathcal{L}_{k}(\boldsymbol{\theta}) \right).
\end{equation}

The further investigation in the work is aimed at studying exactly this difference, which occurs when adding another object to the dataset. We are especially interested in limiting properties when the sample size tends to infinity.

\begin{assumption}\label{assumpt}
    Let $\boldsymbol{\theta}^*$ be the local minimum of both empirical loss functions $\mathcal{L}_{k}(\boldsymbol{\theta})$ and $\mathcal{L}_{k+1}(\boldsymbol{\theta})$, i.e. $\nabla \mathcal{L}_{k}(\boldsymbol{\theta}^*) = \nabla \mathcal{L}_{k+1}(\boldsymbol{\theta}^*) = \mathbf{0}$.
\end{assumption}

This assumption will allow us to investigate the behavior of the loss function landscape, limiting ourselves to considering just one point. 

\subsection{Second-order approximation} 

Let us use second-order Taylor approximation for mentioned above loss functions at $\boldsymbol{\theta}^*$. We suppose that decomposition to the second order will be sufficient to study local behavior. The first-order term vanishes because the gradients $\nabla \mathcal{L}_{k}(\boldsymbol{\theta}^*)$ and $\nabla \mathcal{L}_{k+1}(\boldsymbol{\theta}^*)$ zeroes:
\begin{equation}\label{eq:approx}
    \mathcal{L}_{k}(\boldsymbol{\theta}) \approx \mathcal{L}_{k}(\boldsymbol{\theta}^*) + \dfrac{1}{2} (\boldsymbol{\theta} - \boldsymbol{\theta}^*)\T \mathbf{H}^{(k)}(\boldsymbol{\theta}^*) (\boldsymbol{\theta} - \boldsymbol{\theta}^*),
\end{equation}
where we denoted the Hessian of $\mathcal{L}_{k}(\boldsymbol{\theta})$ w.r.t. parameters $\boldsymbol{\theta}$ at $\boldsymbol{\theta}^*$ as $\mathbf{H}^{(k)}(\boldsymbol{\theta}^*) \in \mathbb{R}^{P \times P}$. Moreover, the total Hessian can be written as the average value of the Hessians of the individual terms of the empirical loss function:
\[ \mathbf{H}^{(k)}(\boldsymbol{\theta}) = \nabla^2_{\boldsymbol{\theta}} \mathcal{L}_{k}(\boldsymbol{\theta}) = \dfrac{1}{k} \sum\limits_{i=1}^{k} \nabla^2_{\boldsymbol{\theta}} \ell(f_{\boldsymbol{\theta}}(\mathbf{x}_{i}), \mathbf{y}_{i}) = \dfrac{1}{k} \sum\limits_{i=1}^{k} \mathbf{H}_{i}(\boldsymbol{\theta}). \]
Therefore, using the obtained second-order approximation~\eqref{eq:approx}, formula for the difference of losses~\eqref{eq:difference} becomes
\[ \mathcal{L}_{k+1}(\boldsymbol{\theta}) - \mathcal{L}_k(\boldsymbol{\theta}) = \dfrac{1}{k+1} \left( \ell(f_{\boldsymbol{\theta}^*}(\mathbf{x}_{k+1}), \mathbf{y}_{k+1}) - \dfrac{1}{k} \sum\limits_{i=1}^{k} \ell(f_{\boldsymbol{\theta}^*}(\mathbf{x}_{i}), \mathbf{y}_{i}) \right) + \]
\[ + \dfrac{1}{k+1} (\boldsymbol{\theta} - \boldsymbol{\theta}^*)\T \left( \mathbf{H}_{k+1}(\boldsymbol{\theta}^*) - \dfrac{1}{k} \sum\limits_{i=1}^{k} \mathbf{H}_{i}(\boldsymbol{\theta}^*) \right) (\boldsymbol{\theta} - \boldsymbol{\theta}^*). \]
After that, using triangle inequality, we can derive the following:
\[ \left| \mathcal{L}_{k+1}(\boldsymbol{\theta}) - \mathcal{L}_k(\boldsymbol{\theta}) \right| \leqslant \dfrac{1}{k+1} \left| \ell(f_{\boldsymbol{\theta}^*}(\mathbf{x}_{k+1}), \mathbf{y}_{k+1}) - \dfrac{1}{k} \sum\limits_{i=1}^{k} \ell(f_{\boldsymbol{\theta}^*}(\mathbf{x}_{i}), \mathbf{y}_{i}) \right| + \]
\[ + \dfrac{1}{k+1} \left\|\boldsymbol{\theta} - \boldsymbol{\theta}^*\right\|_2^2 \left\| \mathbf{H}_{k+1}(\boldsymbol{\theta}^*) - \dfrac{1}{k} \sum\limits_{i=1}^{k} \mathbf{H}_{i}(\boldsymbol{\theta}^*) \right\|_2. \]
So the problem of the boundedness and convergence of the losses difference is reduced to the analysis of the two terms:
\begin{itemize}
    \item Difference of the \textbf{loss functions at optima} for new object and previous ones:
    \[ \left| \ell(f_{\boldsymbol{\theta}^*}(\mathbf{x}_{k+1}), \mathbf{y}_{k+1}) - \dfrac{1}{k} \sum\limits_{i=1}^{k} \ell(f_{\boldsymbol{\theta}^*}(\mathbf{x}_{i}), \mathbf{y}_{i}) \right|, \]
    \item Difference of the \textbf{Hessians at optima} for new object and previous ones:
    \[ \left\| \mathbf{H}_{k+1}(\boldsymbol{\theta}^*) - \dfrac{1}{k} \sum\limits_{i=1}^{k} \mathbf{H}_{i}(\boldsymbol{\theta}^*) \right\|_2. \]
\end{itemize}

It should be mentioned that the first term can be easily upper-bounded by a constant, since the loss function itself takes limited values. However, the expression with Hessians is not so easy to evaluate. The rest of the work is devoted to a thorough analysis of this difference. Thus, we analyze the local convergence of the landscape of the loss function using its Hessian.

\subsection{Fully connected neural network} 

The main results of this work are obtained for $K$-label classification problem with a cross-entropy loss function. In this case the input vector is $\mathbf{x} \in \mathbb{R}^{n}$ and the output $\mathbf{y} \in \mathbb{R}^{K}$, which is a one-hot vector, with all components equal to $0$ except a single component $y_k$ if and only if $k$ is the correct class label for input $\mathbf{x}$. Consider a $L$-layer fully connected network $f_{\boldsymbol{\theta}}(\cdot)$ with ReLU activation function after each linear layer. With $\sigma(\mathbf{x}) = \left[ \mathbf{x} \geqslant \mathbf{0} \right] \mathbf{x}$ as the Rectified Linear Unit (ReLU) function, the output of this network is a vector of logits $\mathbf{z} \in \mathbb{R}^{K}$. They are computed recursively as
\begin{align*}\label{align:fc}
    &\mathbf{z}^{(p)} = \mathbf{W}^{(p)} \mathbf{x}^{(p)} + \mathbf{b}^{(p)}, \\
    &\mathbf{x}^{(p+1)} = \sigma(\mathbf{z}^{(p)}).
\end{align*}
Here we denote the input and output of the $p$-th layer as $\mathbf{x}^{(p)}$ and $\mathbf{z}^{(p)}$, and set $\mathbf{x}^{(1)} = \mathbf{x}$, $\mathbf{z} = f_{\boldsymbol{\theta}}(\mathbf{x}) = \mathbf{z}^{(L)}$. Also we denote $\boldsymbol{\theta} = \mathrm{col}(\mathbf{w}^{(1)}, \mathbf{b}^{(1)}, \ldots, \mathbf{w}^{(L)}, \mathbf{b}^{(L)}) \in \mathbb{R}^{P}$ the parameters of the network. For the $p$-th layer, $\mathbf{w}^{(p)}$ is the flattened weight matrix $\mathbf{W}^{(p)}$ and $\mathbf{b}^{(p)}$ is its corresponding bias vector. We denote $\mathbf{p} = \mathrm{softmax}(\mathbf{z}) \in \mathbb{R}^{K}$ as the output confidence, i.e.
\[ p_i = \mathrm{softmax}(\mathbf{z})_i = \dfrac{\exp{(z_i)}}{\sum_{j=1}^{K} \exp{(z_j)}} \in (0; 1). \] 
The loss function is cross-entropy loss: 
\[ \ell(\mathbf{z}, \mathbf{y}) = \mathrm{CE}(\mathbf{p}, \mathbf{y}) = - \sum_{k=1}^{K} y_k \log p_k \in \mathbb{R}^{+}. \]

\subsection{Hessian decomposition} 

It is quite well known \cite{sagun2018empiricalanalysishessianoverparametrized} that, via the chain rule \cite{skorski2019chainruleshessianhigher}, the Hessian can be decomposed as a sum of the following two matrices: 
\[
    \mathbf{H}_{i}(\boldsymbol{\theta}) = \underbrace{\nabla_{\boldsymbol{\theta}} \mathbf{z}_i \dfrac{\partial^2 \ell(\mathbf{z}_i, \mathbf{y}_i)}{\partial \mathbf{z}_{i}^2} \nabla_{\boldsymbol{\theta}} \mathbf{z}_i\T }_{\text{G-term}} + \underbrace{\sum\limits_{k=1}^{K} \dfrac{\partial \ell(\mathbf{z}_i, \mathbf{y}_i)}{\partial z_{ik}} \nabla^2_{\boldsymbol{\theta}} z_{ik}}_{\text{H-term}},
\]
where $\nabla_{\boldsymbol{\theta}} \mathbf{z}_i \in \mathbb{R}^{P \times K}$ is the Jacobian of the neural network function and $\dfrac{\partial^2 \ell(\mathbf{z}_i, \mathbf{y}_i)}{\partial \mathbf{z}_{i}^2}$ is the Hessian of the loss with respect to the network function, at the $i$-th sample. As it was mentioned \cite{pmlr-v97-ghorbani19b,sagun2018empiricalanalysishessianoverparametrized,papyan2019spectrumdeepnethessiansscale}, the Hessian spectrum consists of  bulk which is concentrated around zero (corresponds to the H-term), and the edges which are scattered away from zero (G-term). We will focus on the top eigenspace, so we can approximate our full Hessian using only G-term, as
\[ \mathbf{H}_{i}(\boldsymbol{\theta}) \approx \nabla_{\boldsymbol{\theta}} \mathbf{z}_i \dfrac{\partial^2 \ell(\mathbf{z}_i, \mathbf{y}_i)}{\partial \mathbf{z}_{i}^2} \nabla_{\boldsymbol{\theta}} \mathbf{z}_i\T. \]

Moreover, recent works exploring the Neural Tangent Kernel \cite{NEURIPS2018_5a4be1fa,Lee_2020} assume that the logits $\mathbf{z}$ depend only linearly on the weights $\boldsymbol{\theta}$, which implies that the logit curvatures $\nabla^2_{\boldsymbol{\theta}} z_{ik}$, and therefore the H-term are identically zero. 

\section{Convergence of the loss difference}\label{sec:converg}

In this section, we present the results obtained regarding the bounding the Hessian norm. After that, we apply the obtained upper bound on the Hessian to get a rate of the losses difference convergence. Despite the fact that this is only an upper bound and it is not always achieved, in the Section~\ref{sec:exp} we analyze the general trends for practical application. 

Authors of \cite{wu2022dissecting} derived the formula of G-term in the fully connected neural network, so in our case we use the following approximation: $\mathbf{H}_{i}(\boldsymbol{\theta}) \approx \mathbf{F}_i\T \mathbf{A}_i \mathbf{F}_i$. Here the following is denoted (we omit the index $i$ for simplicity):
\begin{itemize}
    \item Matrix representation of the ReLU activation function:
    \[ \mathbf{D}^{(p)} = \mathrm{diag}([\mathbf{z}^{(p)} \geqslant \mathbf{0}]), \]
    \item The partial derivative of logits w.r.t. logits at $p$-th layer:
    \[ \mathbf{G}^{(p)} = \dfrac{\partial \mathbf{z}}{\partial \mathbf{z}^{(p)}} = \mathbf{W}^{(L)} \mathbf{D}^{(L-1)} \mathbf{W}^{(L-1)} \mathbf{D}^{(L-2)} \cdot \ldots \cdot \mathbf{D}^{(p)}, \]
    \item Its stacked version:
    \[ \mathbf{F}\T = \begin{pmatrix}
        (\mathbf{G}^{(1)})\T \otimes \mathbf{x}^{(1)} \\
        (\mathbf{G}^{(1)})\T \\ 
        \vdots \\
        (\mathbf{G}^{(L)})\T \otimes \mathbf{x}^{(L)} \\
        (\mathbf{G}^{(L)})\T \\ 
    \end{pmatrix}, \]
    \item And the Hessian of the loss function w.r.t. logits (according to \cite{singla2019understanding}):
    \[ \mathbf{A} = \nabla^2_\mathbf{z} \ell(\mathbf{z}, \mathbf{y}) = \mathrm{diag}(\mathbf{p}) - \mathbf{p} \mathbf{p}\T. \]
\end{itemize}

To the best of our knowledge, no one had received an estimate for the norms of such an approximation of the Hessian previously. Next, we present a Theorem~\ref{thm:hess} that contains an upper bound on the spectral norm of the Hessian in a fully connected neural network.

\subsection{Boundedness of the Hessian}

Below there is a Theorem~\ref{thm:hess}, a detailed proof of which is provided in Appendix~\ref{app:thm:hess}. 

\begin{theorem}\label{thm:hess}
    Consider a $L$-layer fully connected neural network with ReLU activation function and without bias terms, applied to solve a $K$-label classification problem. Suppose the following is satisfied: $\| \mathbf{W}^{(p)} \|_2 \leqslant M_{\mathbf{W}}$ and $\| \mathbf{x}_i \|_2 \leqslant M_{\mathbf{x}}$ for all layers $p = 1, \ldots, L$ in network and for all objects $i = 1, \ldots, m$ in the dataset. Then, for any object $i = 1, \ldots, m$ the following inequality holds:
    \[ \left\| \mathbf{H}_i(\boldsymbol{\theta}) \right\|_2 \leqslant L \sqrt{2} M_{\mathbf{x}}^2 M_{\mathbf{W}}^{2L} + \sqrt{2} \dfrac{M_{\mathbf{W}}^2 (M_{\mathbf{W}}^{2L} - 1)}{M_{\mathbf{W}}^2 - 1}. \]
\end{theorem}

This theorem allows us to understand the dependence of the Hessian norm on the structure of the neural network: the size of the hidden layer and the number of layers. So, the next Lemma~\ref{lemma:hess} exhibits the dependence on the hidden size.

\begin{lemma}\label{lemma:hess}
    If each model parameter is bounded by a constant $M > 0$, that is $|w_{ij}^{(p)}| \leqslant M$ for all $i, j = 1, \ldots, h$ and for all layers $p = 1, \ldots, L$, then, under the conditions of Theorem~\ref{thm:hess}, the following is true:
    \[ \left\| \mathbf{H}_i(\boldsymbol{\theta}) \right\|_2 \leqslant L \sqrt{2} M_{\mathbf{x}}^2 (hM)^{2L} + \sqrt{2} \dfrac{(hM)^2 ((hM)^{2L} - 1)}{(hM)^2 - 1}. \]
    So, the following proportionality holds:
    \[ \left\| \mathbf{H}_i(\boldsymbol{\theta}) \right\|_2 \propto L (hM)^{2L}. \]
\end{lemma}

Obtained results allows us to claim that the Hessian norm is a power function of the size of the hidden layer $h$ and an exponential function of the number of layers $L$. Although it may seem that the estimate received is too high, this is actually not the case. The fact is that if we choose $h$ to be large, then the limiting constant $M$ will most likely be very small. Because of this, the number under the power of $2L$ will probably be less than one. Further, we use this upper bound to get an inequality for the loss function difference.

\subsection{Losses difference convergence}\label{subseq:rate}

Below there is a Lemma~\ref{lemma:loss}, a detailed proof of which is provided in Appendix~\ref{app:lemma:loss}

\begin{lemma}\label{lemma:loss}
    Let $\boldsymbol{\theta}$ be chosen as $\left\|\boldsymbol{\theta} - \boldsymbol{\theta}^*\right\|_2^2 \leqslant R^2$ for some $R > 0$. If there exist a non-negative constant $M_{\ell}$ such $\left| \ell(f_{\boldsymbol{\theta}^*}(\mathbf{x}_{i}), \mathbf{y}_{i}) \right| \leqslant M_{\ell}$ for all objects $i = 1, \ldots, m$ in the dataset, then, under the conditions of Theorem~\ref{thm:hess}, the following holds:
    \[ \left| \mathcal{L}_{k+1}(\boldsymbol{\theta}) - \mathcal{L}_k(\boldsymbol{\theta}) \right| \leqslant \dfrac{2}{k+1}\left( M_{\ell} + \left( L \sqrt{2} M_{\mathbf{x}}^2 M_{\mathbf{W}}^{2L} + \sqrt{2} \dfrac{M_{\mathbf{W}}^2 (M_{\mathbf{W}}^{2L} - 1)}{M_{\mathbf{W}}^2 - 1} \right) R^2 \right) \to 0 \text{ as } k \to \infty. \]
    So, the following proportionality is true:
    \begin{equation}\label{eq:rate}
        \left| \mathcal{L}_{k+1}(\boldsymbol{\theta}) - \mathcal{L}_k(\boldsymbol{\theta}) \right| \propto \dfrac{L (hM)^{2L} R^2}{k}. 
    \end{equation}
\end{lemma}

Based on the estimates obtained, the following conclusions can be drawn. Firstly, the convergence of the loss function surface is affected by the distance from the extremum point. The farther we are from this point, the slower the convergence may be. Secondly, the situation regarding the number of layers and layer size is similar to that of the Hessian matrix. An increase in the number of layers $L$ will lead to a higher convergence score, while an increase in layer size $h$ does not necessarily have a negative impact. Again, this is due to the constant that evaluates the magnitude of each element in the weight matrix. Finally, the resulting estimate is inversely proportional to the sample size, exhibiting a sublinear rate of convergence. In the next section, we demonstrate the typical behavior of the loss function surface in practice and compare it with the theoretical one obtained.

\section{Experiments}\label{sec:exp}

To verify the theoretical estimates obtained, we conducted a detailed empirical study. In this section, we present the results from training a fully connected neural network for the \textbf{Image Classification} task. The experiments can be easily reproduced by following the instructions provided in our GitHub repository: \href{https://github.com/kisnikser/landscape-hessian}{https://github.com/kisnikser/landscape-hessian}.

The primary objective of these experiments is to empirically confirm the convergence of the loss landscape as the sample size increases. To achieve this, we trained a fully connected neural network on \textbf{the entire dataset} and obtained the corresponding parameters $\hat{\boldsymbol{\theta}}$ as a point near the minimum. Subsequently, we examined the relationship between the average loss difference and the available sample size.

We utilized the \texttt{pytorch} library \cite{pytorch} as the Python framework for neural network training. The architecture employed is consistent with that described in Section~\ref{sec:prelim}, consisting of several linear layers with a ReLU activation function after each layer, except the final one. The size $h$ was fixed for all hidden layers $L$. The network was trained over numerous epochs using the \texttt{Adam} optimizer \cite{adam} with a constant learning rate of $10^{-3}$. Various Image Classification datasets available in the \texttt{torchvision} library were used. While the graphs presented below are based on a single dataset, additional results for each dataset can be found in Appendix~\ref{app:exp}. For the training process, a batch size of 64 was selected. The experiments were conducted on a Tesla A100 80GB GPU with 16 CPU cores and 243 GB of RAM.

\subsection{Direct Image Classification}

\begin{figure}[ht]
    \centering
    \includegraphics[width=0.5\linewidth]{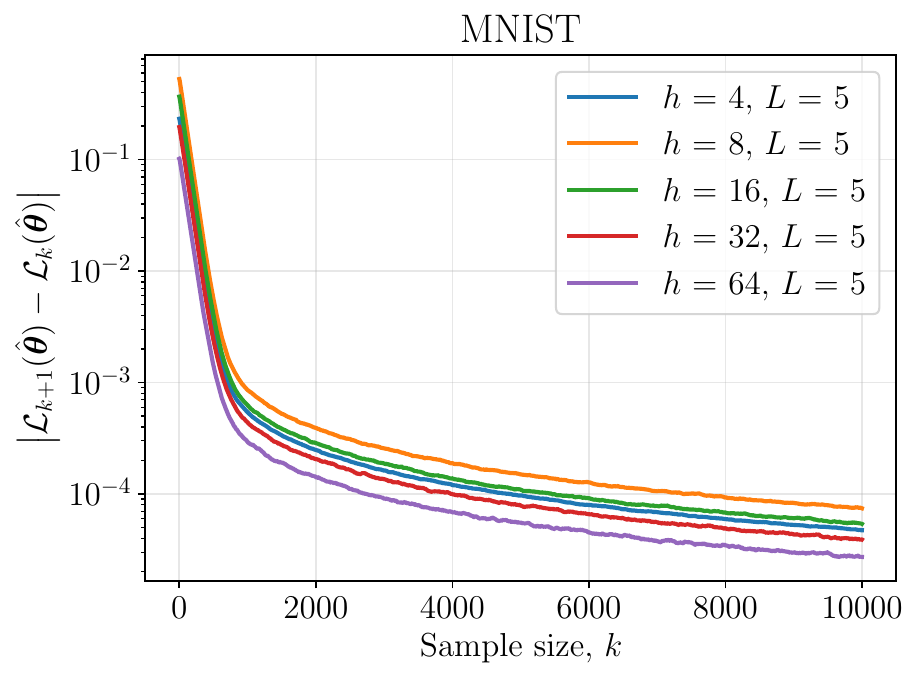}\hfill
    \includegraphics[width=0.5\linewidth]{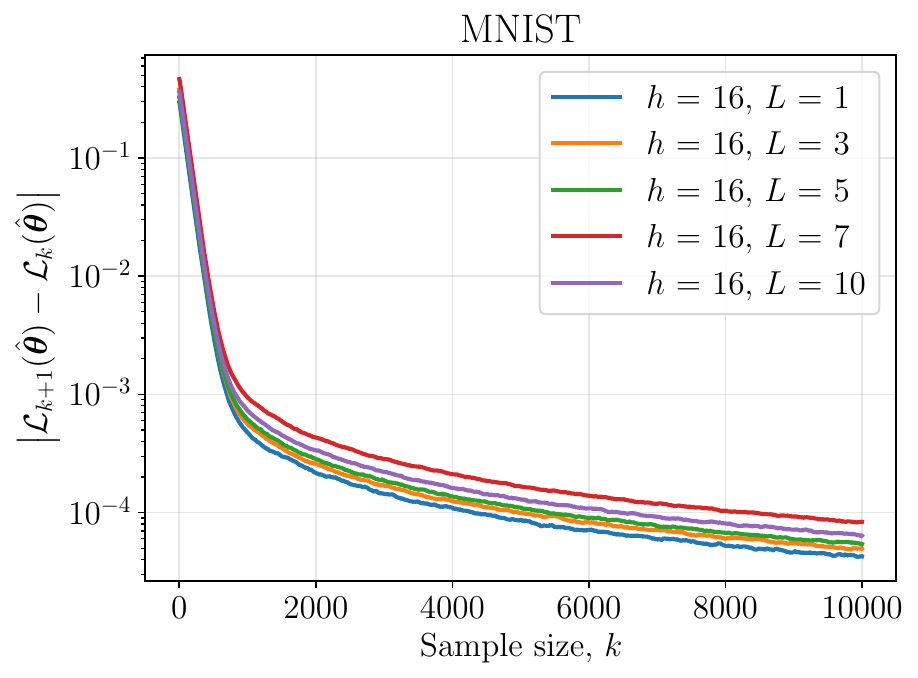}
    \caption{The dependence of the absolute value of the loss function difference on the available sample size, \textbf{direct image classification}. The graphs on the left show a decrease in values as the dimension of the hidden layer increases. The graphs on the right show an increase in values as the number of layers increases.}
    \label{fig:mnist}
\end{figure}

In this experiment, we utilized the pixel values of images as inputs. Figure~\ref{fig:mnist} displays the results obtained from an analysis of 10,000 objects from the MNIST dataset \cite{deng2012mnist}, with the network trained over 10 epochs. The corresponding input size is 784, while the output size is 10. The plots on the left were generated by fixing the number of layers at $L=5$ in the network. The hidden size across all layers was varied from 4 to 64. Concurrently, the figure on the right illustrates the behavior of the loss difference as the number of hidden layers changes from 1 to 10, with the hidden size $h$ fixed at 16. This sequence was repeated 100 times for averaging. An exponential moving average with a smoothing factor of 0.99 was applied to the obtained results.

From the dependencies observed, it is evident that although the change is not substantial, adding more layers leads to a greater difference in the loss functions (see the \textbf{right} side). Conversely, increasing the hidden size results in a smaller difference between the loss functions (see the \textbf{left} side). 

For readers unfamiliar with the topic, this may be surprising, as the estimation we have made \eqref{eq:rate} suggests a power dependence on the layer size $h$, and the value of $L$ is clearly not less than 1. However, readers are referred to Section~\ref{subseq:rate} for a more detailed discussion of this phenomenon. Additionally, in practice, the constant $M$ that limits the magnitude of weights is found to be relatively small. Furthermore, since the MNIST dataset classification task is considered relatively straightforward, a shallow yet wide neural network can produce good classification results. Consequently, the values of the loss function have been observed to be lower for larger $h$ values, and therefore their difference is also lower.

\subsection{Image features extraction} 

In contrast to the previous experiment, this part employs a pre-trained image feature extractor. The fully connected network is utilized as a multi-label classification head. We selected the Vision Transformer (ViT) \cite{wu2020visual} from Google.

\begin{figure}[ht]
    \centering
    \includegraphics[width=0.5\linewidth]{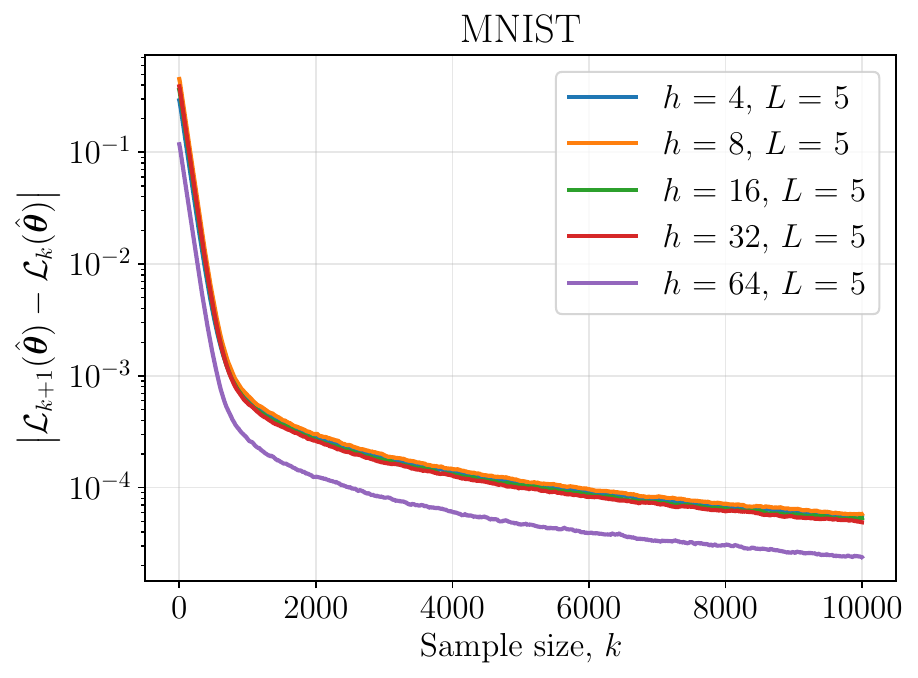}\hfill
    \includegraphics[width=0.5\linewidth]{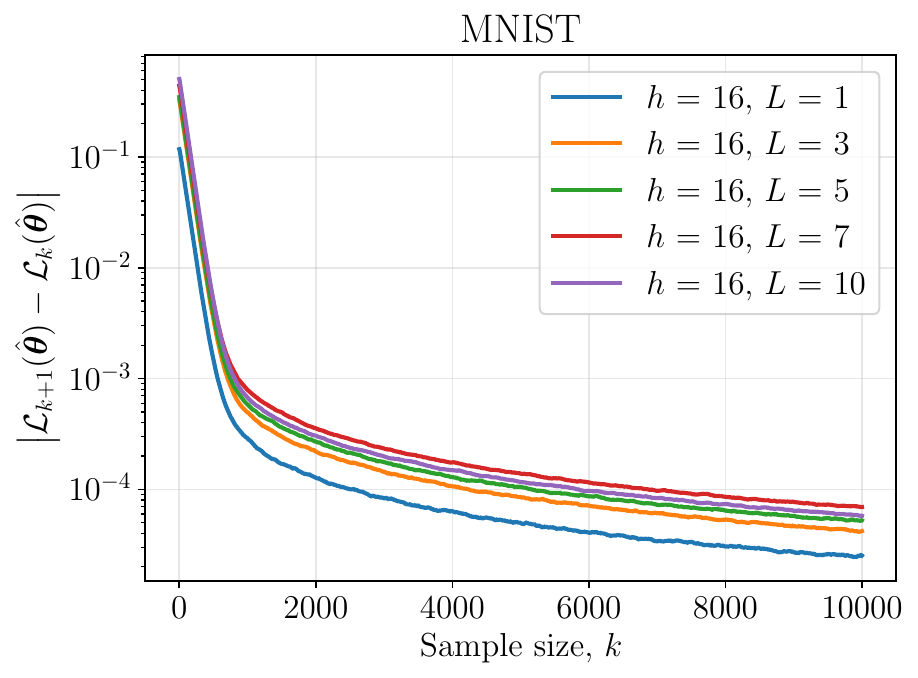}
    \caption{The dependence of the absolute value of the loss function difference on the available sample size, \textbf{image features extraction}. The graphs on the left show a decrease in values as the dimension of the hidden layer increases. The graphs on the right show an increase in values as the number of layers increases.}
    \label{fig:mnist-extraction}
\end{figure}

We similarly selected 10,000 objects randomly from the MNIST dataset and varied the hidden size and the number of layers. The results are consistent with those observed in direct image classification. This consistency confirms that the convergence presented does not depend on the nature of the space of the original objects $\mathcal{X}$. Boundedness of this space is sufficient to observe the convergence of the loss function landscape.

Our experiment corroborates the convergence proved in Lemma~\ref{lemma:loss}. Additionally, the upper bound on the rate of this convergence holds true. Indeed, altering the parameters of the neural network, such as the number of layers and the layer size, leads to a slight change in the difference of the loss functions. We remind the reader that a larger number of graphs can be found in Appendix~\ref{app:exp}.

\section{Discussion}\label{sec:disc}

The results of this study provide insights into the loss landscape convergence as the dataset size increases. Our theoretical analysis shows that the absolute difference between the average loss function values when adding one more object to the sample tends to zero, as the number of available objects tends to infinity. This was achieved by proving the upper-bound theorem for the Hessian norm in a fully connected neural network. Empirical study allows us to confirm our results practically. In particular, we claim that the loss function surface exhibits convergence for the Image Classification task, both as a direct classifier of initial representations and as a multi-label classification head after the pre-trained feature extractor. 

The results of our research are highly connected to the problem of the local geometry of neural loss landscapes. Despite the fact that a large number of studies have been devoted to this issue, the change in dataset size has remained a significant gap. In this paper, we have tried to take the first steps in this direction. 

Nevertheless, this study has potential limitations. First, our theoretical analysis was deterministic and not probabilistic. Although this may bring certain clarifications to the estimates, we do not think it can have a serious impact in practice. Second, the subject of our research is a fully connected neural network. We plan to extend our results to other architectures in future work. Third, using Assumption~\ref{assumpt}, we suppose existing such a point, which will be a minimum, starting with a certain sample size. Finally, using the triangle inequality in the proof of Lemma~\ref{lemma:loss} yields a rough upper bound for the loss difference, so it may be improved in future work. 

We believe that our findings will contribute to the development of more precise studies of the behavior of the loss landscape when changing the training sample size. We also believe that our results can be used to develop modern sample-size determination techniques. We expect this because the convergence of the loss function landscape can be considered as the sign that the training sample size is sufficient for the selected model. In future work, we hope to apply our observations to this field.

\section{Conclusion}\label{sec:concl}

In this paper, we have presented a comprehensive study of the convergence of the loss landscape in a fully connected neural network as the sample size increases. Our theoretical analysis and empirical results demonstrate that the absolute difference between the average loss function values when adding one more object to the sample tends to zero as the number of available objects tends to infinity. These findings provide valuable insights into the local geometry of neural loss landscapes and address a previously unexplored issue in the field. We believe that our results will contribute to the development of more precise studies of the loss landscape behavior and have implications for the development of sample size determination techniques. Future work will focus on extending our results to other architectures and improving the upper bounds for the loss difference.


\bibliographystyle{unsrtnat}
\bibliography{references}


\newpage
\appendix
\section{Appendix / supplemental material}\label{app}

\subsection{Additional experiments}\label{app:exp}

In this section, we provide an extended version of the conducted experiments. As it was discussed in the Section~\ref{sec:exp}, we trained a fully connected neural network for the Image Classification task. Below there is a Table~\ref{table:datasets} with a description of the datasets used. We have chosen four datasets from the \texttt{torchivison} library: MNIST~\cite{deng2012mnist}, FashionMNIST~\cite{xiao2017fashionmnistnovelimagedataset}, CIFAR10 and CIFAR100~\cite{krizhevsky2009learning}. The only preprocessing of the data is normalization to bring the values into the range $[-1; 1]$.

\begin{table}[ht]
  \caption{Image Classification datasets description}
  \label{table:datasets}
  \centering
  \begin{tabular}{llll}
    \toprule
    Name     & Description     & Format & Resolution \\
    \midrule
    MNIST \cite{deng2012mnist} & Handwritten digits & Grayscale & $28 \times 28$ \\
    FashionMNIST \cite{xiao2017fashionmnistnovelimagedataset} & Fashion clothing items & Grayscale & $28 \times 28$ \\
    CIFAR10 \cite{krizhevsky2009learning} & Various objects & RGB & $32\times 32$ \\
    CIFAR100 \cite{krizhevsky2009learning} & Various objects & RGB & $32 \times 32$ \\
    \bottomrule
  \end{tabular}
\end{table}

We have already discussed the results for the MNIST dataset in Section~\ref{sec:exp}, so the following plots are only for the remained sets.

\textbf{Direct image classification.} Here we used pixel values of images as inputs. Plots on the left were gained when fixing the number of layers $L=5$ in the network. We changed the hidden size on the all layers from $4$ to $64$. At the same time, the figure on the right demonstrates the behavior of the loss difference when the number of hidden layers changes from $1$ to $10$, but the size $h = 16$ remains unchanged. This sequence was repeated $100$ times for averaging. For the obtained results, we applied an exponential moving average with smoothing factor $0.99$.

\begin{figure}[ht] 
  \begin{subfigure}[b]{0.5\linewidth}
    \centering
    \includegraphics[width=\linewidth]{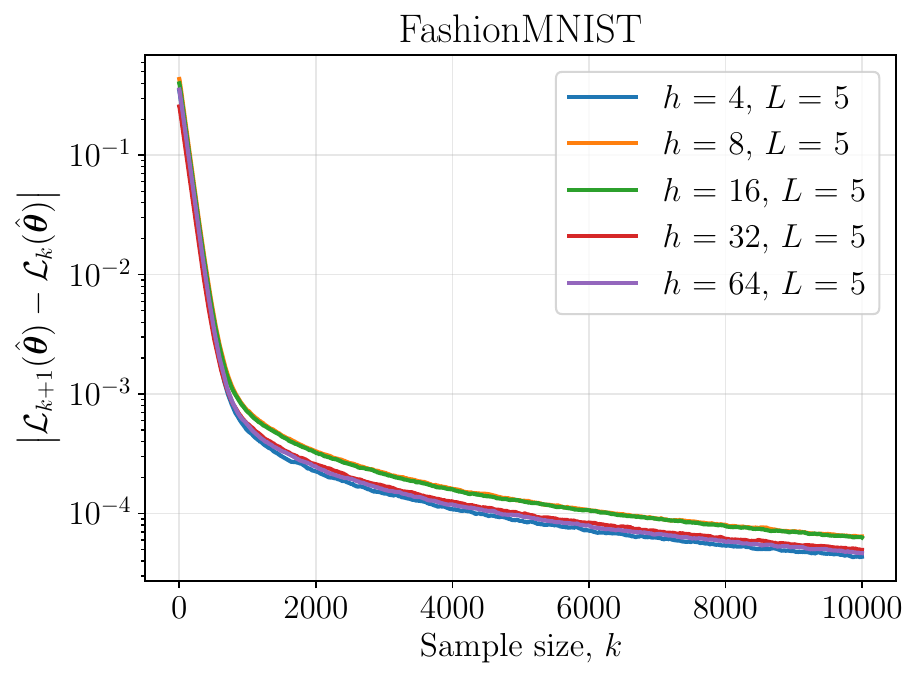} 
  \end{subfigure}
  \begin{subfigure}[b]{0.5\linewidth}
    \centering
    \includegraphics[width=\linewidth]{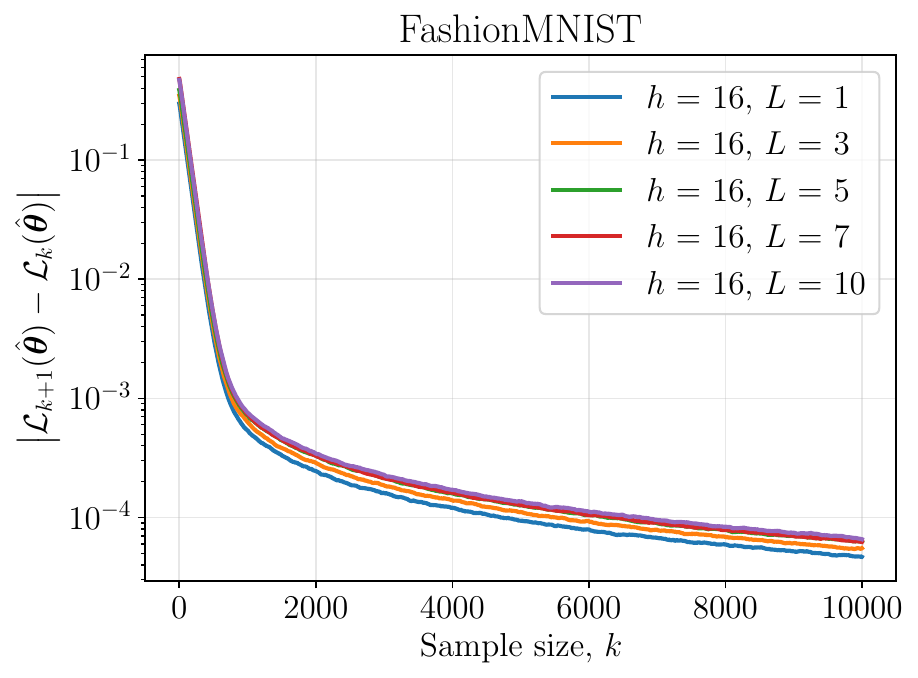} 
  \end{subfigure} 
  \begin{subfigure}[b]{0.5\linewidth}
    \centering
    \includegraphics[width=\linewidth]{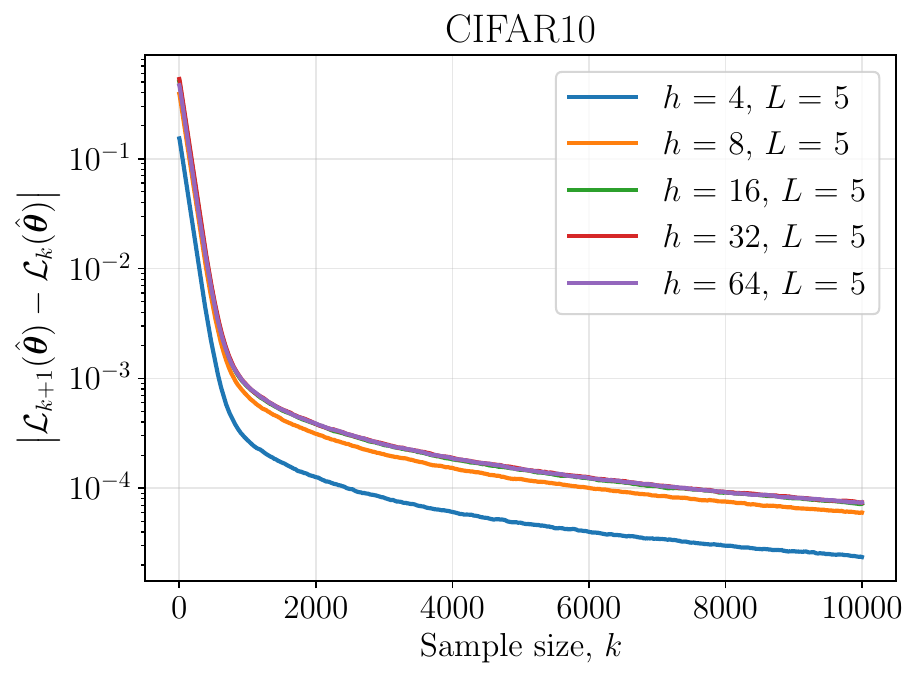} 
  \end{subfigure}
  \begin{subfigure}[b]{0.5\linewidth}
    \centering
    \includegraphics[width=\linewidth]{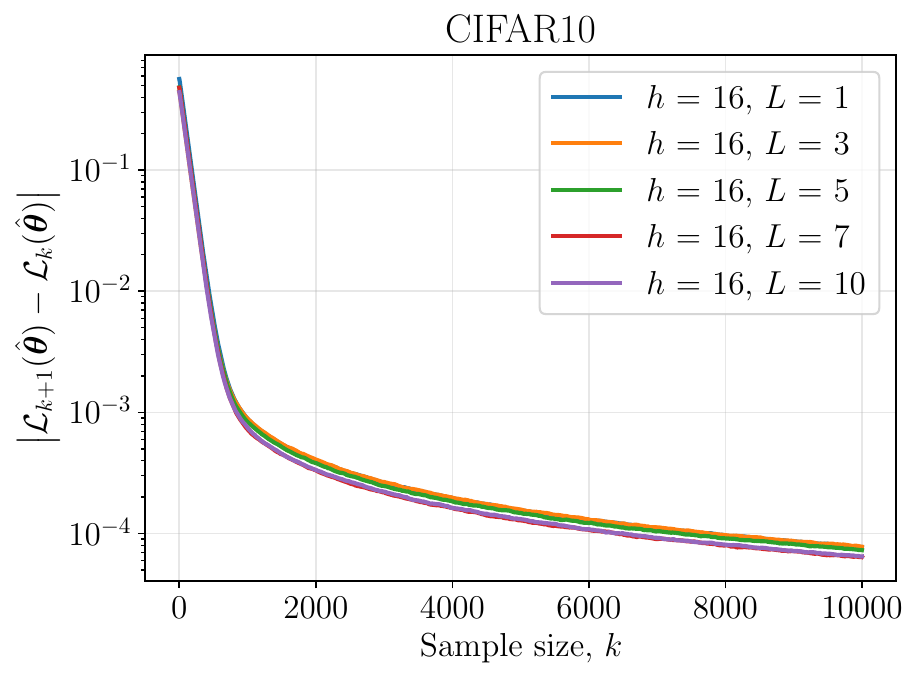} 
  \end{subfigure} 
  \begin{subfigure}[b]{0.5\linewidth}
    \centering
    \includegraphics[width=\linewidth]{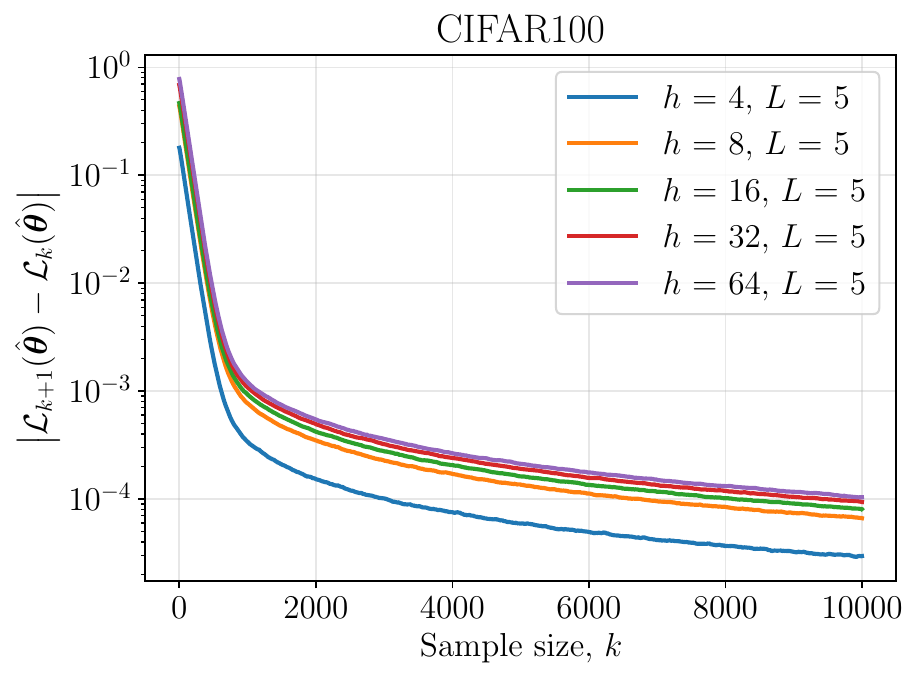} 
  \end{subfigure}
  \begin{subfigure}[b]{0.5\linewidth}
    \centering
    \includegraphics[width=\linewidth]{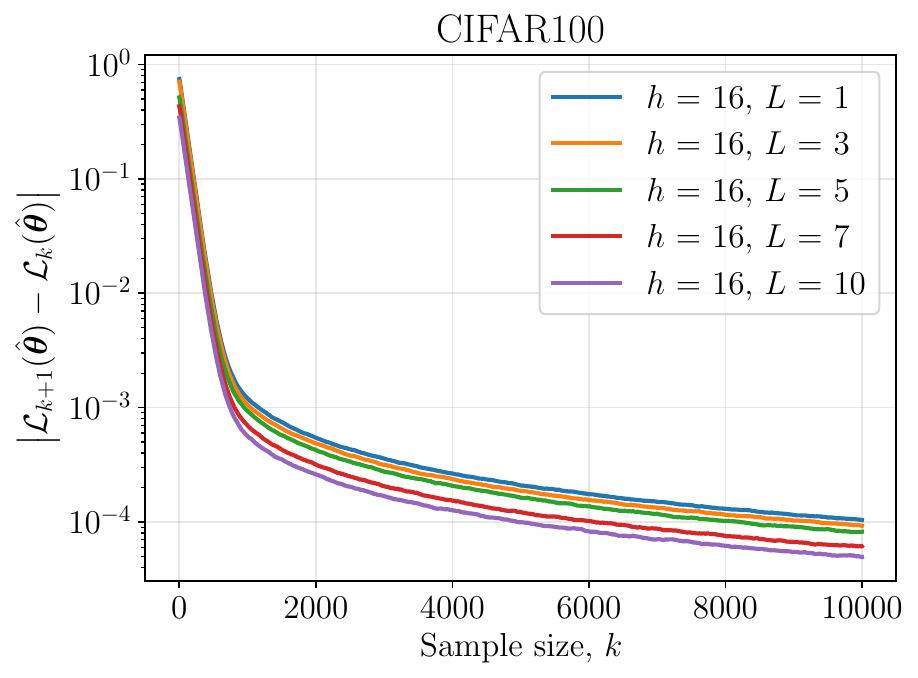} 
  \end{subfigure} 
  \caption{The dependence of the absolute value of the loss function difference on the available sample size, \textbf{direct image classification}. The graphs on the left show a decrease in values as the dimension of the hidden layer increases. The graphs on the right show an increase in values as the number of layers increases. Results on different datasets: FashionMNIST, CIFAR10 and CIFAR100.}
  \label{fig:additional-exp} 
\end{figure}

\textbf{Image features extraction.} Unlike the previous experiment part, here we use a pre-trained image feature extractor firstly. 

\begin{figure}[ht]
  \begin{subfigure}[b]{0.5\linewidth}
    \centering
    \includegraphics[width=\linewidth]{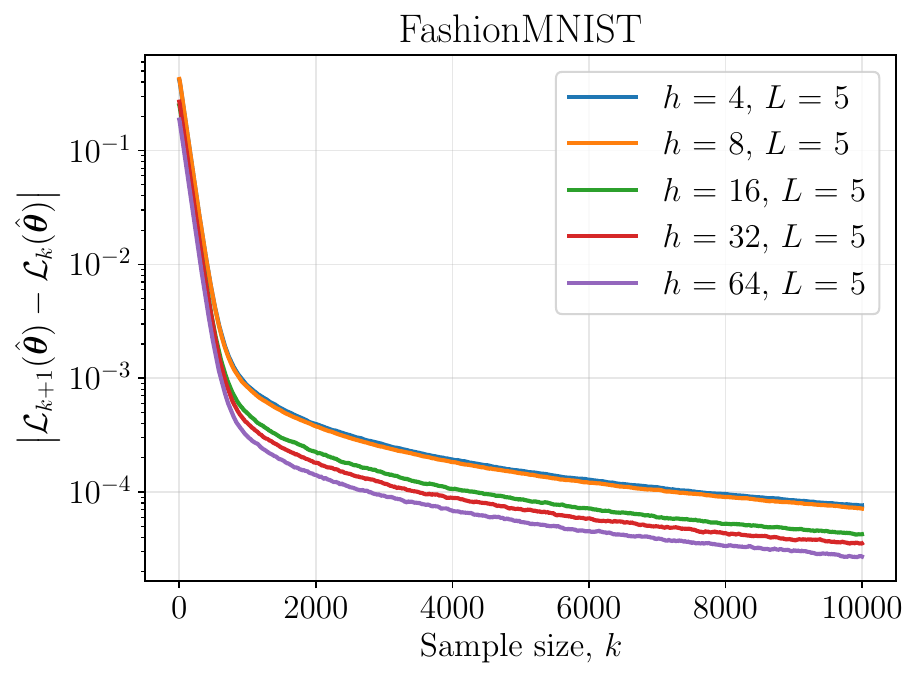} 
  \end{subfigure}
  \begin{subfigure}[b]{0.5\linewidth}
    \centering
    \includegraphics[width=\linewidth]{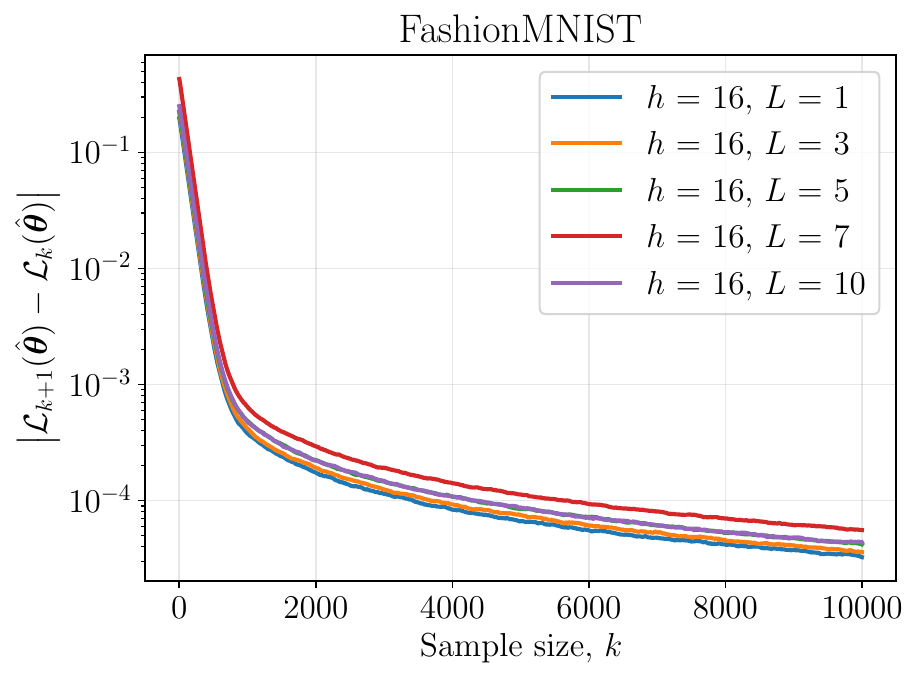} 
  \end{subfigure} 
  \begin{subfigure}[b]{0.5\linewidth}
    \centering
    \includegraphics[width=\linewidth]{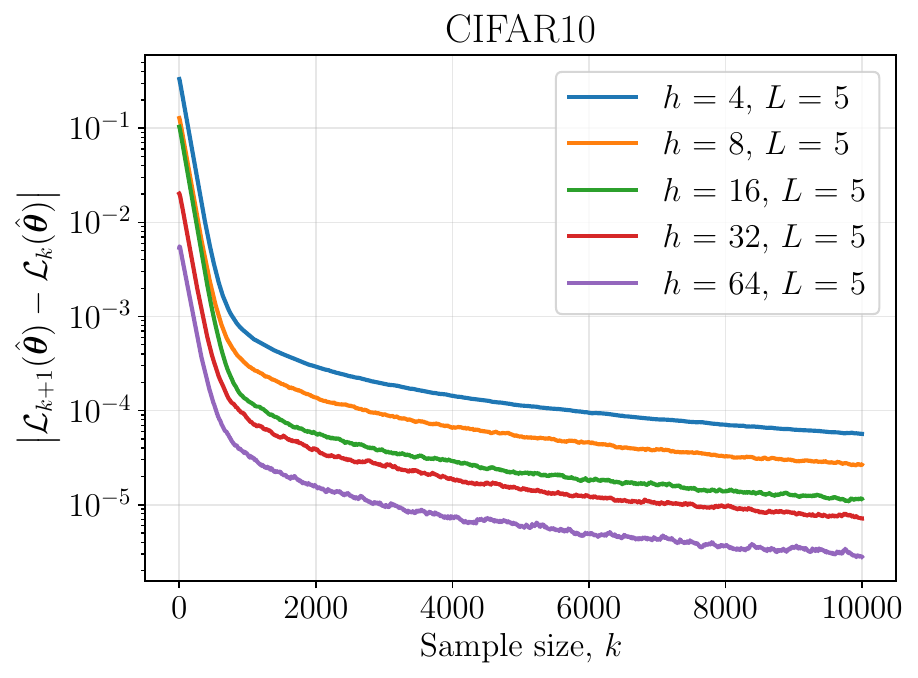} 
  \end{subfigure}
  \begin{subfigure}[b]{0.5\linewidth}
    \centering
    \includegraphics[width=\linewidth]{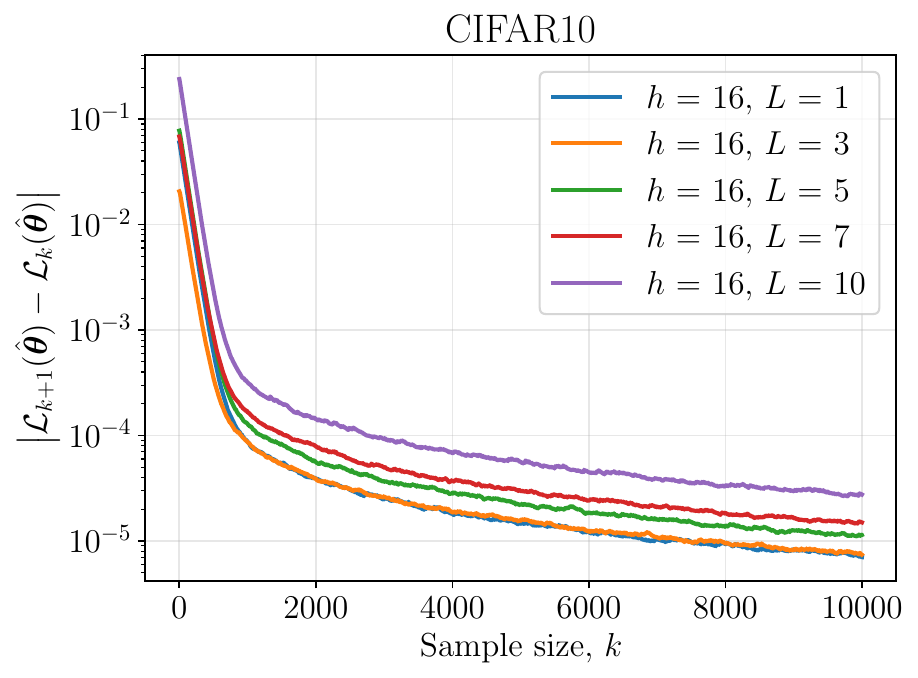} 
  \end{subfigure} 
  \begin{subfigure}[b]{0.5\linewidth}
    \centering
    \includegraphics[width=\linewidth]{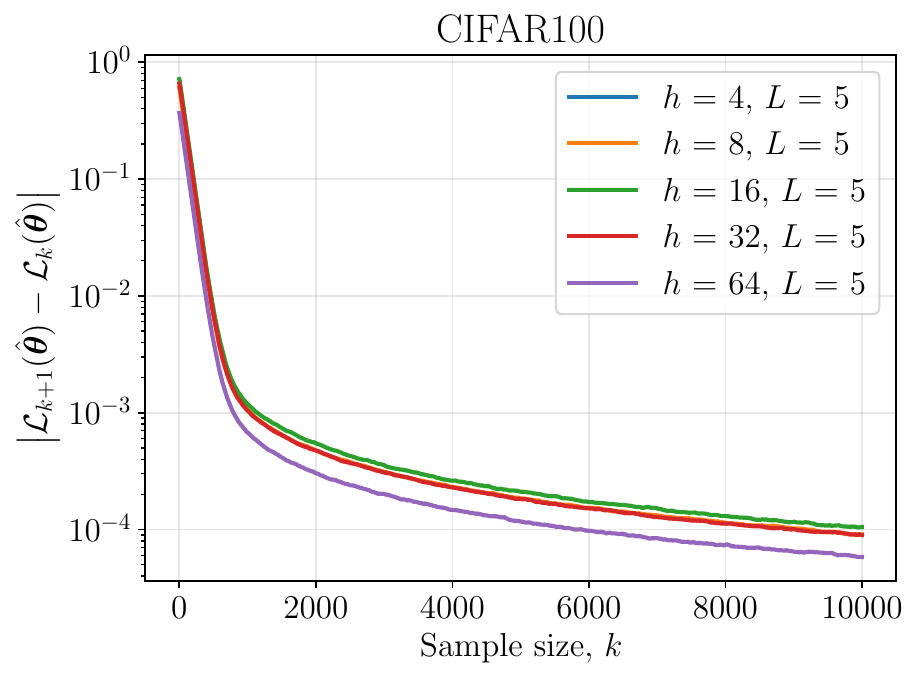} 
  \end{subfigure}
  \begin{subfigure}[b]{0.5\linewidth}
    \centering
    \includegraphics[width=\linewidth]{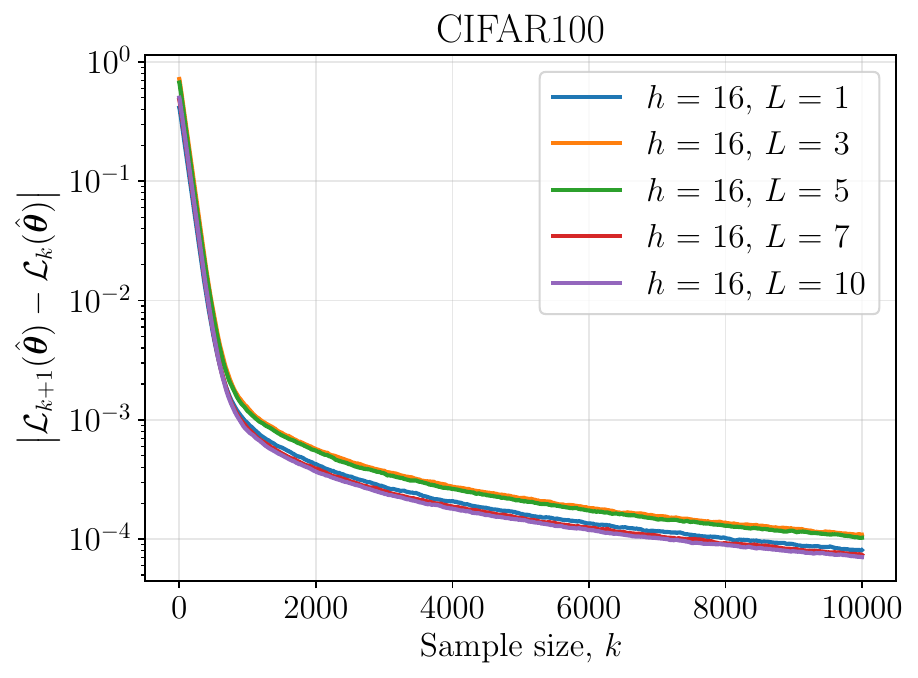} 
  \end{subfigure} 
  \caption{The dependence of the absolute value of the loss function difference on the available sample size, \textbf{image features extraction}. The graphs on the left show a decrease in values as the dimension of the hidden layer increases. The graphs on the right show an increase in values as the number of layers increases. Results on different datasets: FashionMNIST, CIFAR10 and CIFAR100.}
  \label{fig:additional-exp-extraction} 
\end{figure}

Similarly to Section~\ref{sec:exp}, the results confirm the convergence proved in the Lemma~\ref{lemma:loss}. Also, the upper bound on the rate of this convergence is true. Specifically, changing the parameters of the neural network: the number of layers and the size of the layer leads to the change in the difference of the loss functions.

\clearpage
\newpage

\subsection{Proof of Theorem~\ref{thm:hess}}\label{app:thm:hess}
\begin{proof}
    For simplicity, we omit the index $i$, which corresponds to the particular object in dataset.
    Firstly, because the spectral matrix norm is sub-multiplicative, we get
    \[ \left\| \mathbf{H}(\boldsymbol{\theta}) \right\|_2 = \left\| \mathbf{F}\T \mathbf{A} \mathbf{F} \right\|_2 \leqslant \left\| \mathbf{A} \right\|_2 \left\| \mathbf{F} \right\|_2^2. \]
    Next, we will focus on considering each term separately.
    $\left\| \mathbf{A} \right\|_2$:
    Due to the norm equivalence, the following inequality holds:
    \[ \left\| \mathbf{A} \right\|_2 \leqslant \left\| \mathbf{A} \right\|_F, \]
    where $\left\| \mathbf{A} \right\|_F$ is the Frobenius norm. So, further we will use some key properties of this norm to estimate the above term. Using the definition of Frobenius norm,
    \[ \left\| \mathbf{A} \right\|_F^2 = \left\| \mathrm{diag}(\mathbf{p}) - \mathbf{p} \mathbf{p}\T \right\|_F^2 = \sum\limits_{k=1}^{K} (p_k - p_k^2)^2 + \sum\limits_{k \neq l} p_k^2 p_l^2 = \sum\limits_{k=1}^{K} p_k^2 (1 - p_k)^2 + \sum\limits_{k \neq l} p_k^2 p_l^2. \]
    Since $0 \leqslant p_k \leqslant 1$ for all $i = 1, \ldots, K$, we have $0 \leqslant p_k^2 \leqslant p_k$ and $0 \leqslant (1 - p_k)^2 \leqslant (1 - p_k)$. Consequently, we can derive the following inequalities:
    \[ \sum\limits_{k=1}^{K} p_k^2 (1 - p_k)^2 \leqslant \sum\limits_{k=1}^{K} p_k (1 - p_k) \leqslant \sum\limits_{k=1}^{K} p_k = 1, \]
    \[ \sum\limits_{k \neq l} p_k^2 p_l^2 \leqslant \sum\limits_{k \neq l} p_k p_l = \left( \sum\limits_{k=1}^{K} p_k \right)^2 - \sum\limits_{k=1}^{K} p_k^2 \leqslant 1 - \sum\limits_{k=1}^{K} p_k^2. \]
    Combining these inequalities, we obtain:
    \[ \left\| \mathbf{A} \right\|_F^2 \leqslant 1 + 1 - \sum\limits_{k=1}^{K} p_k^2 = 2 - \sum\limits_{k=1}^{K} p_k^2 \leqslant 2, \]
    thus, the Frobenius norm of $\mathbf{A}$ is upper-bounded by $\sqrt{2}$, therefore
    \[ \left\| \mathbf{A} \right\|_2 \leqslant \left\| \mathbf{A} \right\|_F \leqslant \sqrt{2}. \]
    $\left\| \mathbf{F} \right\|_2$:
    To obtain a bound on the $\left\| \mathbf{F} \right\|_2$, we firstly analyze the spectral norm of the matrix 
    \[ \mathbf{G}^{(p)} = \mathbf{W}^{(L)} \mathbf{D}^{(L-1)} \mathbf{W}^{(L-1)} \mathbf{D}^{(L-2)} \cdot \ldots \cdot \mathbf{D}^{(p)}.\] 
    Using the sub-multiplicative property of the spectral norm, we have:
    \[ \| \mathbf{G}^{(p)} \|_2 \leqslant \|\mathbf{W}^{(L)}\|_2 \|\mathbf{D}^{(L-1)}\|_2 \|\mathbf{W}^{(L-1)}\|_2 \|\mathbf{D}^{(L-2)}\|_2 \cdot \ldots \cdot \|\mathbf{D}^{(p)}\|_2. \]
    Since $\mathbf{D}^{(p)}$ is a diagonal matrix with entries equal to 0 or 1, its spectral norm is upper-bounded by 1. Therefore, we can simplify the above inequality as:
    \[ \| \mathbf{G}^{(p)} \|_2 \leqslant \prod_{s=p}^{L} \| \mathbf{W}^{(s)} \|_2. \]
    Then, using the property that squared spectral norm of vertically-stacked matrices is less or equal to the sum of their squared spectral norms (it is easy to observe), we get:
    \[ \| \mathbf{F} \|_2^2 \leqslant \sum\limits_{p=1}^{L} \left( \| (\mathbf{G}^{(1)})\T \otimes \mathbf{x}^{(1)} \|_2^2 + \| (\mathbf{G}^{(1)})\T \|_2^2 \right). \]
    Spectral norm of the Kronecker matrix product is equal to their ordinary product norm, i.e.
    \[ \| \mathbf{F} \|_2^2 \leqslant \sum\limits_{p=1}^{L} \| \mathbf{G}^{(p)} \|_2^2 \left(\| \mathbf{x}^{(p)} \|_2^2 + 1\right). \]
    Further, we substitute the upper-bound obtained above and have:
    \[ \| \mathbf{F} \|_2^2 \leqslant \sum\limits_{p=1}^{L} \left(\| \mathbf{x}^{(p)} \|_2^2 + 1\right) \prod_{s=p}^{L} \| \mathbf{W}^{(s)} \|_2^2 . \]
    So the final bound we get for the Hessian is (we substitute the object index again):
    \[ \left\| \mathbf{H}_i(\boldsymbol{\theta}) \right\|_2 \leqslant \sqrt{2} \sum\limits_{p=1}^{L} \left(\| \mathbf{x}_i^{(p)} \|_2^2 + 1\right) \prod_{s=p}^{L} \| \mathbf{W}^{(s)} \|_2^2 . \]
    To the simplicity, we will omit bias terms, i.e. set $\mathbf{b}^{(p)} = \mathbf{0}$ for all $p = 1, \ldots, L$, then we get the following:
    \[ \| \mathbf{x}_i^{(p)} \|_2 \leqslant \| \mathbf{x}_i \|_2 \prod_{s=1}^{p-1} \| \mathbf{W}^{(s)} \|_2, \]
    and therefore
    \[ \left\| \mathbf{H}_i(\boldsymbol{\theta}) \right\|_2 \leqslant \sqrt{2} \sum\limits_{p=1}^{L} \left( \| \mathbf{x}_i \|_2^2 \prod_{s=1}^{p-1} \| \mathbf{W}^{(s)} \|_2^2 + 1\right) \prod_{s=p}^{L} \| \mathbf{W}^{(s)} \|_2^2 = \]
    \[ = L \sqrt{2} \| \mathbf{x}_i \|_2^2 \prod_{p=1}^{L} \| \mathbf{W}^{(p)} \|_2^2 + \sqrt{2} \sum\limits_{p=1}^{L} \prod_{s=p}^{L} \| \mathbf{W}^{(s)} \|_2^2. \]
    If $\| \mathbf{W}^{(p)} \|_2 \leqslant M_{\mathbf{W}}$ and $\| \mathbf{x}_i \|_2 \leqslant M_{\mathbf{x}}$, then
        \[ \left\| \mathbf{H}_i(\boldsymbol{\theta}) \right\|_2 \leqslant L \sqrt{2} M_{\mathbf{x}}^2 M_{\mathbf{W}}^{2L} + \sqrt{2} \dfrac{M_{\mathbf{W}}^2 (M_{\mathbf{W}}^{2L} - 1)}{M_{\mathbf{W}}^2 - 1}. \]
    A separate interesting case is when $L = 1$, then
    \[ \left\| \mathbf{H}_i(\boldsymbol{\theta}) \right\|_2 \leqslant \sqrt{2} M_{\mathbf{W}}^{2} \left( M_{\mathbf{x}}^2 + 1 \right) . \]
\end{proof}

\subsection{Proof of Lemma~\ref{lemma:hess}}\label{app:lemma:hess}
\begin{proof}
    The relationship between the spectral norm and the Frobenius norm, as well as the definition of the Frobenius norm, allow us to obtain
    \[ \| \mathbf{W}^{(p)} \|_2^2 \leqslant \| \mathbf{W}^{(p)} \|_F^2 = \sum_{i, j = 1}^{h} \left(w_{ij}^{(p)}\right)^2 \leqslant h^2 M^2, \]
    where $M$ is such a constant, that $\left(w_{ij}^{(p)}\right)^2 \leqslant  M^2$ for all $i, j = 1, \ldots, h$ and for all $p = 1, \ldots, L$.
    \[ h^2 M^2 \leqslant M_{\mathbf{W}}^2, \]
    therefore
    \[ \left\| \mathbf{H}_i(\boldsymbol{\theta}) \right\|_2 \leqslant L \sqrt{2} M_{\mathbf{x}}^2 (hM)^{2L} + \sqrt{2} \dfrac{(hM)^2 ((hM)^{2L} - 1)}{(hM)^2 - 1}. \]
    So, the following proportionality is true:
    \[ \left\| \mathbf{H}_i(\boldsymbol{\theta}) \right\|_2 \propto L h^{2L}. \]
    
    A separate interesting case is when $L = 1$, then
    \[ \left\| \mathbf{H}_i(\boldsymbol{\theta}) \right\|_2 \leqslant \sqrt{2} M_{\mathbf{W}}^{2} \left( M_{\mathbf{x}}^2 + 1 \right), \]
    that is
    \[ \left\| \mathbf{H}_i(\boldsymbol{\theta}) \right\|_2 \propto h^2. \]
\end{proof}

\subsection{Proof of Lemma~\ref{lemma:loss}}\label{app:lemma:loss}
\begin{proof}
    Using the triangle inequality for the mentioned above terms, we get
    \begin{align*}
        & \left| \ell(f_{\boldsymbol{\theta}^*}(\mathbf{x}_{k+1}), \mathbf{y}_{k+1}) - \dfrac{1}{k} \sum\limits_{i=1}^{k} \ell(f_{\boldsymbol{\theta}^*}(\mathbf{x}_{i}), \mathbf{y}_{i}) \right| \leqslant \\
        & \leqslant \left| \ell(f_{\boldsymbol{\theta}^*}(\mathbf{x}_{k+1}), \mathbf{y}_{k+1}) \right| + \left| \dfrac{1}{k} \sum\limits_{i=1}^{k} \ell(f_{\boldsymbol{\theta}^*}(\mathbf{x}_{i}), \mathbf{y}_{i}) \right| \leqslant \\
        & \leqslant \left| \ell(f_{\boldsymbol{\theta}^*}(\mathbf{x}_{k+1}), \mathbf{y}_{k+1}) \right| + \dfrac{1}{k} \sum\limits_{i=1}^{k} \left| \ell(f_{\boldsymbol{\theta}^*}(\mathbf{x}_{i}), \mathbf{y}_{i}) \right| \leqslant
    \end{align*}
    Then, due to the boundedness of loss on train samples, we obtain
    \[ \leqslant M_{\ell} + \dfrac{1}{k} \sum\limits_{i=1}^{k} M_{\ell} = 2M_{\ell} = \mathcal{O}(1) \text{ as } k \to \infty. \]
    Similarly for Hessians, it is easy to get
    \begin{align*}
        & \left\| \mathbf{H}_{k+1}(\boldsymbol{\theta}^*) - \dfrac{1}{k} \sum\limits_{i=1}^{k} \mathbf{H}_{i}(\boldsymbol{\theta}^*) \right\|_2 \leqslant \\
        & \leqslant \left\| \mathbf{H}_{k+1}(\boldsymbol{\theta}^*) \right\| + \left\| \dfrac{1}{k} \sum\limits_{i=1}^{k} \mathbf{H}_{i}(\boldsymbol{\theta}^*) \right\|_2 \leqslant \\
        & \leqslant \left\| \mathbf{H}_{k+1}(\boldsymbol{\theta}^*) \right\| + \dfrac{1}{k} \sum\limits_{i=1}^{k} \left\| \mathbf{H}_{i}(\boldsymbol{\theta}^*) \right\|_2 \leqslant \\
        & \leqslant M_{\mathbf{H}} + \dfrac{1}{k} \sum\limits_{i=1}^{k} M_{\mathbf{H}} = 2 M_{\mathbf{H}} = \mathcal{O}(1) \text{ as } k \to \infty,
    \end{align*}
    where from Theorem~\ref{thm:hess} we get
    \[ M_{\mathbf{H}} = L \sqrt{2} M_{\mathbf{x}}^2 M_{\mathbf{W}}^{2L} + \sqrt{2} \dfrac{M_{\mathbf{W}}^2 (M_{\mathbf{W}}^{2L} - 1)}{M_{\mathbf{W}}^2 - 1}. \]
    
    Thus, substituting the obtained estimates into the expression for the difference, we receive
    \[ \left| \mathcal{L}_{k+1}(\boldsymbol{\theta}) - \mathcal{L}_k(\boldsymbol{\theta}) \right| \leqslant \dfrac{2M_{\ell}}{k+1} + \dfrac{2M_{\mathbf{H}}}{k+1} \left\|\boldsymbol{\theta} - \boldsymbol{\theta}^*\right\|_2^2. \]
    Choosing the particular neighborhood of the local minimum $\boldsymbol{\theta}^*$, i.e. $\left\|\boldsymbol{\theta} - \boldsymbol{\theta}^*\right\|_2^2 \leqslant R^2$, we get
    \[ \left| \mathcal{L}_{k+1}(\boldsymbol{\theta}) - \mathcal{L}_k(\boldsymbol{\theta}) \right| \leqslant \dfrac{2}{k+1}\left( M_{\ell} + M_{\mathbf{H}}R^2 \right) \to 0 \text{ as } k \to \infty. \]
\end{proof}

\end{document}